%% file: paper.tex
\documentclass[letterpaper, 10 pt, conference]{ieeeconf}  

\IEEEoverridecommandlockouts                              

\usepackage{amsmath, amsfonts} 

\usepackage{algorithm}
\usepackage{algpseudocode}
\usepackage{graphicx}
\usepackage{textcomp}
\usepackage{xcolor}
\usepackage{gensymb}
\usepackage{forloop}
\usepackage{flushend}
\usepackage{hyperref}
\hypersetup{
    colorlinks=true,
    linkcolor=blue,
    filecolor=magenta,      
    urlcolor=cyan,
    pdftitle={Overleaf Example},
    pdfpagemode=FullScreen,
    }

\usepackage{eso-pic}
\newcommand\AtPageUpperMyright[1]{\AtPageUpperLeft{
 \put(\LenToUnit{0.1\paperwidth},\LenToUnit{-1cm}){
     \parbox{0.9\textwidth}{\raggedleft\fontsize{9}{11}\selectfont #1}}
 }}
\newcommand{\conf}[1]{
\AddToShipoutPictureBG*{
\AtPageUpperMyright{#1}
}
}

\conf{Author final version of the manuscript accepted for publication in the 31st IEEE International Conference on Robot \& Human Interactive Communication (RO-MAN 2022), \textcopyright IEEE, \textbf{DOI}:\href{https://doi.org/10.1109/RO-MAN53752.2022.9900740}{10.1109/RO-MAN53752.2022.9900740}} 

\usepackage[noadjust]{cite}

\algblockdefx{When}{EndWhen}[1]{\textbf{when} #1}{\textbf{end}}
\algblockdefx{Every}{EndEvery}[1]{\textbf{every} #1}{\textbf{end}}
\algblockdefx{In}{EndIn}[1]{\textbf{In} #1}{\textbf{end}}

\algnewcommand\algorithmicforeach{\textbf{for each}}
\algdef{S}[FOR]{ForEach}[1]{\algorithmicforeach\ #1\ \algorithmicdo}

\algnewcommand\algorithmicswitch{\textbf{switch}}
\algnewcommand\algorithmiccase{\textbf{case}}
\algnewcommand\algorithmicupdateplan{\texttt{updatePlan}}
\algnewcommand\updatePlan[1]{\State \algorithmicupdateplan(#1)}%
\algdef{SE}[SWITCH]{Switch}{EndSwitch}[1]{\algorithmicswitch\ #1\ \algorithmicdo}{\algorithmicend\ \algorithmicswitch}%
\algdef{SE}[CASE]{Case}{EndCase}[1]{\algorithmiccase\ #1}{\algorithmicend\ \algorithmiccase}%
\algtext*{EndSwitch}%
\algtext*{EndCase}%

\title{\LARGE \bf Knowing Where to Look: A Planning-based Architecture \\to Automate the Gaze Behavior of Social Robots*}

\author{Chinmaya Mishra$^{1}$ and Gabriel Skantze$^{2}$
\thanks{*This project has received funding from the European Union's Framework Programme for Research and Innovation Horizon 2020 (2014-2020) under the Marie Sklodowska-Curie Grant Agreement No. 859588.}
\thanks{$^{1}$Chinmaya Mishra is an Associate Researcher and PhD Fellow at Furhat Robotics, Stockholm.
        {\tt\small chinmaya@furhatrobotics.com}}%
\thanks{$^{2}$Gabriel Skantze is a Professor at the Department of Speech Music and Hearing at KTH Royal Institute of Technology, Stockholm and Co-founder at Furhat Robotics, Stockholm.
        {\tt\small skantze@kth.se}}%
}

\usepackage{tikz}
\usepackage{textcomp}
\usepackage{hyperref}
\usepackage{lipsum}

\newcommand\copyrighttext{%
  \footnotesize \textcopyright 2022 IEEE. Personal use of this material is permitted.
  Permission from IEEE must be obtained for all other uses, in any current or future
  media, including reprinting/republishing this material for advertising or promotional
  purposes, creating new collective works, for resale or redistribution to servers or
  lists, or reuse of any copyrighted component of this work in other works.}
\newcommand\copyrightnotice{%
\begin{tikzpicture}[remember picture,overlay]
\node[anchor=south,yshift=10pt] at (current page.south) {\fbox{\parbox{\dimexpr\textwidth-\fboxsep-\fboxrule\relax}{\copyrighttext}}};
\end{tikzpicture}%
}

\begin{document}



\maketitle
\copyrightnotice

\thispagestyle{empty}
\pagestyle{empty}

\begin{abstract}
Gaze cues play an important role in human communication and are used to coordinate turn-taking and joint attention, as well as to regulate intimacy. In order to have fluent conversations with people, social robots need to exhibit human-like gaze behavior. Previous Gaze Control Systems (GCS) in HRI have  automated robot gaze using data-driven or heuristic approaches. However, these systems tend to be mainly reactive in nature. Planning the robot gaze ahead of time could help in achieving more realistic gaze behavior and better eye-head coordination. In this paper, we propose and implement a novel planning-based GCS. We evaluate our system in a comparative within-subjects user study (N=26) between a reactive system and our proposed system. The results show that the users preferred the proposed system and that it was significantly more interpretable and better at regulating intimacy. 
\end{abstract}


\section{Introduction}\label{intro}
Non-verbal cues play a crucial role in realizing effective communication in Human-Human Interaction (HHI). Humans make use of many non-verbal cues such as eye gaze, facial expressions, gestures, and prosody to convey meaning during social interactions. Among these non-verbal cues, eye gaze cues are considered to be especially important, as they are interpreted using dedicated and unique hard-wired pathways in the brain~\cite{emery2000eyes}. 
Interpreting and conveying feelings and intentions through eye gaze during a social interaction is central to HHI and comes naturally to humans. During social interactions, gaze cues are used to coordinate turn-taking~\cite{kendon1967some}, signal cognitive effort~\cite{argyle1976gaze}, and regulate intimacy~\cite{abele1986functions}, among other things. 

As social robots become increasingly available in society, they are expected to be able to communicate using both verbal and non-verbal cues, similar to humans. Research has shown that the robot's gaze behavior plays a similarly important role in Human-Robot Interaction (HRI)~\cite{yamazaki2008precision,imai2002robot}. Consequently, researchers have  designed architectures to control the gaze behaviors of robots to explore the impact of social gaze in HRI and exploit the many uses of gaze cues in social interactions \cite{admoni2017social, pereira2019responsive}. Most of these Gaze Control Systems (GCS) have generally focused on modelling specific gaze cues such as gaze aversion \cite{andrist2014conversational} or turn-taking \cite{mutlu2012conversational}. 

Even though these GCS have achieved good results in emulating human-like gaze behaviors in robots, a common limitation is that they remain mainly \textit{reactive} in nature. Although some of these systems do plan the gaze behavior for the upcoming utterance at the onset of the utterance (e.g. \cite{andrist2014conversational}), the plan does not get updated incrementally, and the plan does not really affect the gaze behavior in the current moment. 
Another limitation of many systems is that they are static, in the sense that they use fixed durations for gaze shifts. For example, in \cite{pereira2019responsive}, the gaze of the robot was fixed on the relevant target for a duration of 1-5 seconds during the interaction, before moving to the target with the lowest priority. 
In contrast, HHI involves a lot of \textit{planning}. Research has shown that gaze behavior is coordinated with the underlying speech plan \cite{BEATTIE2010}. Depending on how long we plan to look at something, we determine whether a quick glance would suffice or whether we need to move the head and look. In this study, we focus on bringing a planning component into GCS. 
More specifically, we address the following research question:

\textit{How can planning be used to generate better gaze behavior in HRI?}

Our model plans the priority for each potential gaze target (e.g., users or objects) in the environment incrementally (frame-by-frame) for a future rolling time window. At each time step, various events in the conversation might update these priorities, resulting in an evolving gaze plan which produces gaze behavior that is dynamic and differs in frequency and duration based on the state of the conversation. This planning allows the robot to better coordinate eye and head movements, since it is possible to compute for how long the robot will be attending a specific target in advance. In addition, the robot can better plan when to avert the gaze to regulate intimacy. The model is comprehensive in that it encompasses turn-taking, gaze aversion (GA), referential gaze (RG) and responsive joint attention (RJA). We evaluate the proposed model and compare it with a purely reactive heuristic model, using a multi-party interaction scenario where the robot head Furhat collaborates with two human players to sort a deck of cards in the right order. 

The main contributions of this paper are: 
\begin{itemize}
    \item A comprehensive gaze control architecture that accounts for turn-taking, joint attention and intimacy regulation in HRI, using a planning-based approach. 
    \item A novel approach to make use of the planned gaze behaviors to coordinate eye-head movements of the robot. 
    \item An evaluation done in a complex multi-party HRI setting, which shows that this system is better than a reactive version of the same system. 
\end{itemize}

\section{Related Work}\label{relatedWorks}
Modelling approaches in HRI for GCS can be broadly categorized into \textit{data-driven} approaches (e.g., \cite{andrist2014conversational,mutlu2012conversational}), where HHI gaze data are used to build models that can predict the gaze  of the robot, and \textit{heuristic} approaches, where the gaze of the robot is controlled using a set of rules, based on findings from HHI literature (e.g., \cite{pereira2019responsive,mehlmann2014exploring}). 

\textbf{Turn-taking} refers to the process in which interlocutors coordinate and take turns while speaking \cite{Skantze2021, jokinen2013gaze}. \cite{mutlu2012conversational} implemented data-driven models for role-signalling, turn-taking and topic signalling gaze mechanisms based on the formal observations of human communication. 
It was found that the subjects were able to correctly interpret the turn-yielding signals by the robot 99\% of the time. 

\textbf{Gaze aversion} is the intentional shifting of the gaze away from the interaction partner during a conversation. Several studies have focused on modelling this and evaluate human perception of it. \cite{andrist2014conversational} lists three primary functions of gaze aversion: cognitive, intimacy regulation and turn-taking. 
They used human gaze aversion data to model gaze aversion on a NAO robot and found that gaze aversion by the robot was perceived to be intentional. 
\cite{zhong2019investigating} implemented a GCS with four possible states to model mutual gaze and gaze aversion using the captured gaze data of the participants. 
\cite{lala2019smooth} used a heuristic model to generate appropriate gaze aversion along with verbal fillers as turn-taking cues. 

When the interlocutors attend to a common target during a social interaction (and are mutually aware of that), it is generally referred to as \textbf{joint attention}. Joint attention is usually split into two categories: \textit{responding to joint attention} (\textbf{RJA}) and \textit{initiating joint attention} (\textbf{IJA}) \cite{mundy2007attention}. IJA refers to when the interlocutor initiates a joint attention by directing the gaze at the referent (also known as \textbf{referential gaze}). RJA refers to the act of following others' gaze direction and interpreting the need to share focus on a common point.  \cite{mehlmann2014exploring} proposed and implemented a GCS (\textit{Sceneflow}) that made use of the bi-directional and multimodal aspects of speech. The model implemented referential gaze, RJA and mutual gaze as a hierarchical and concurrent state-chart-based architecture. 
\cite{pereira2019responsive} focused on the effects that RJA has on people's perception of social robots. The GCS was divided into two layers: \textit{Proactive Gaze Layer} and \textit{Responsive Gaze Layer} which modelled RJA and IJA respectively with each module having a predefined priority used to suppress gaze shifts issued by other modules with lower priorities. 

Several studies have also developed models of human gaze behavior which could then be transferred to robots. \cite{stefanov2019modeling} used a supervised learning approach to predict eye gaze direction or head orientation of the participant in multi-party open world dialogues. 
A recent study modelled the robot's gaze behavior using concepts from animation instead of grounding it in human psychomotor behavior~\cite{pan2020realistic}. 

Even though data-driven approaches are potentially able to provide a more accurate representation of human gaze behavior as compared to heuristic models, they are restricted by their dependence on collecting appropriate gaze data. It is also unclear how well they generalize to settings different from that in which the data was recorded. Another problem is that the speakers' intentions are not available in the data, which makes it difficult for data-driven models to account for planning. 

Another aspect of gaze modelling is the coordination between eye and head movements during gaze shifts. Previous studies have found that human eye and head movements are coordinated based on the target angles to realize gaze shifts \cite{uemura1980eye, stahl1999amplitude}. \cite{hendrikse2018influence} defined eye-head angle relationships 
to control the eye-head movements of a virtual avatar during gaze shifts. \cite{gu2006gaze} and \cite{wijayasinghe2019head} tried to model realistic eye-head coordination on humanoid robots. To the best of our knowledge, ours is the first work that incorporates a planning component into a GCS to coordinate the eye-head movement during gaze shifts and regulate intimacy during a conversation.

\input{gazeModels}

We summarize these previous works on modelling GCS in Table \ref{tab:gazeModels}. Planning here refers to the ability of a GCS to make use of the planned gaze behavior and the executed gaze behavior to adjust the current gaze behavior. As can be seen, our proposed model is unique in its comprehensiveness and its use of planning to control the gaze behavior of the robot.

\section{Test-bed: Card Game}\label{cardGame}
\begin{figure}[htbp] 
  \centering
  \includegraphics[width=\linewidth]{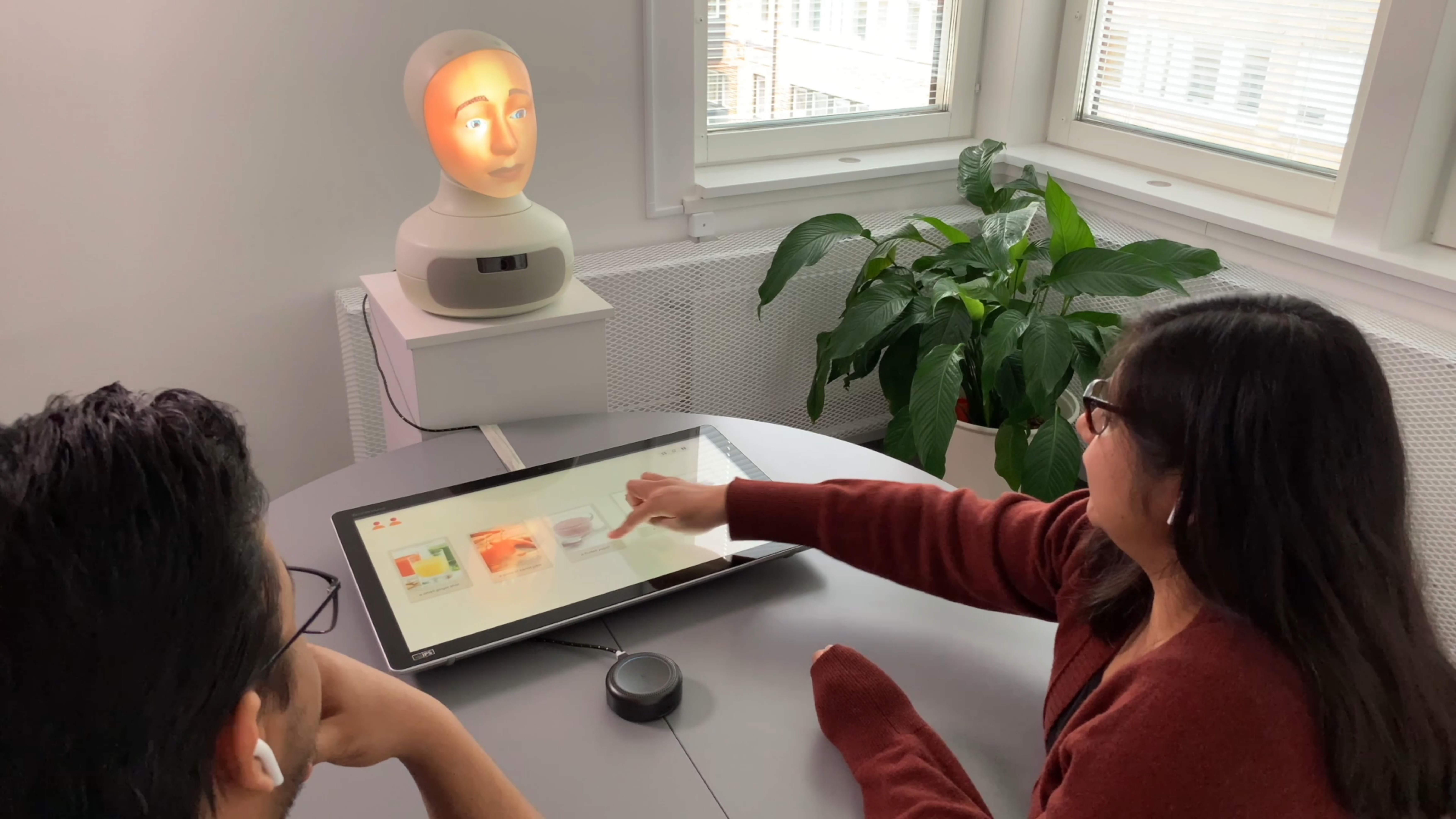}
  \caption{Third person view of the Card Game setup}
  \label{cardGameSetup}
\end{figure}

Our GCS was tested with the \textit{Card Game} scenario, which is a test-bed specifically designed for studying multi-party interactions involving joint attention to objects \cite{skantze2015collaborative}. The Card Game setup consists of a Furhat robot~\cite{moubayed2013furhat}, a touchscreen and up to two players, as seen in Fig.~\ref{cardGameSetup}. A set of cards are shown on the touchscreen and the task is to sort the cards based on some criterion. For example, the task could be to order a set of animals from slowest to fastest based on their running speeds. Furhat and the players then collaborate with each other to arrange the cards in the right order by moving them on the touchscreen.  

During the game, players are encouraged to discuss among each other and with Furhat to reach a solution. Furhat's arguments are based on a randomized belief model, which means that the players have to choose whether they trust Furhat's beliefs or not.
This results in a fairly free form of multi-party conversation, and therefore constitutes a good test-bed for studying turn-taking, joint attention and gaze aversions.
When players look at or move a card, Furhat can display RJA behavior and when Furhat talks about the cards or the game, it can generate referential gaze.

\section{A Comprehensive Gaze Control Architecture}
\begin{figure}[t] 
\centering
  \begin{center}
  \includegraphics[width=\linewidth]{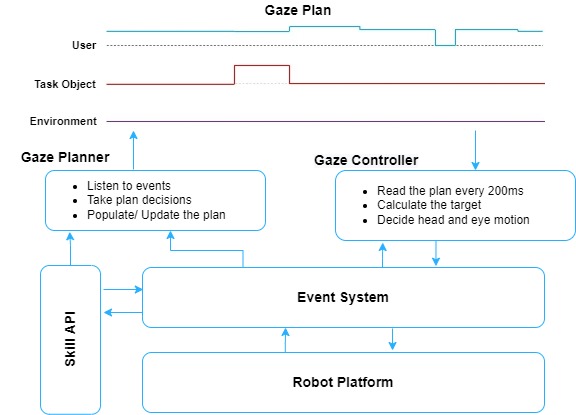}    
  \end{center}
  \caption{Overview of the proposed Gaze Control System}
  \label{gazeArchitecture}
\end{figure}
 Fig.~\ref{gazeArchitecture} shows the overall architecture of the GCS and how it is integrated.
The \textit{Robot Platform} consists of Furhat's output interfaces (projector, neck servo motors, etc.), input devices (microphone, camera, touchscreen, etc.), as well as the software modules for automatic speech recognition (ASR), text-to-speech synthesis (TTS), face tracking, etc. All the sensory inputs, modules and actuators in the Robot Platform are mediated by the \textit{Event System}. The \textit{Gaze Planner} subscribes to high-level events, such as the position of the user, speech input, and location of objects on the touchscreen, to generate a \textit{Gaze Plan}. This plan is then used by the \textit{Gaze Controller}, to generate events that make the robot move the eyes and turn the head. Interactions are implemented using the \textit{Skill API}, with which all the interaction specific details are defined, such as the utterances of the robot, its facial expressions and head gestures, among others. 

\algtext*{EndWhen}
\algtext*{EndFor}
\algtext*{EndIn}
\algtext*{EndWhile}

\begin{algorithm}[t]
\caption{Outline of the Gaze Control System}\label{alg:gazeControlSystem}
\begin{algorithmic}
\ForEach{time step}:
\In{\textbf{Gaze Planner:}}
\State  updateTargets($GP$) \Comment{Add/Remove $\mathcal{T}_i$}
\ForEach{new event $E$}
\When{$E$ is \textit{RobotSpeaking}}
    \State checkPauses($GP$) \Comment{see \ref{sec:planning-turn-taking}}
    \State checkTurnYielding($GP$) \Comment{see \ref{sec:planning-turn-taking}}
    \State checkReferentialGaze($GP$) \Comment{see \ref{sec:jointAttention}}
\EndWhen
\When{$E$ is \textit{TargetsMoved}}
    \State attendTarget($GP$) \Comment{see \ref{sec:jointAttention}}
\EndWhen
\When{$E$ is \textit{UserSpeaking}}
    \State checkRJA($GP$) \Comment{see \ref{sec:jointAttention}}
    \State attendSpeaker($GP$)  \Comment{see \ref{sec:planning-turn-taking}}
\EndWhen
\When{$E$ is \textit{RobotListening}}
    \State attendUser($GP$) \Comment{see \ref{sec:planning-turn-taking}}
\EndWhen
\EndFor
\State checkIntimacyRegulation($GP$) \Comment{see \ref{sec:intimacyRegulation}}

\EndIn
\In{\textbf{Gaze Controller:}} \Comment{see \ref{sec:eyeHeadCoordination}}
    \State $GP_c$ = summarize($GP$)
    \State $\mathcal{T}_c, slack$ = getTarget($GP_c$) 
    \State headAngle = getHeadAngle($\mathcal{T}_c, slack$)
    \State setRobotEyes($\mathcal{T}_c$)
    \State setRobotNeck(headAngle)
\EndIn
\State shift($GP$)
\EndFor
\end{algorithmic}
\end{algorithm}

Algorithm \ref{alg:gazeControlSystem} provides an overview of how the GCS works. During the course of an interaction, the Gaze Planner identifies and maintains a set of \textit{gaze targets} ($\mathcal{T}_i$, $i \in [1\mathinner {\ldotp \ldotp}n]$) along with their current locations in real-time. These gaze targets could be of different types, such as \textit{users}, \textit{task objects} or the \textit{environment}. The GCS continuously monitors the gaze targets to add new targets or remove targets as necessary. 

We introduce a priority score $\mathcal{P} \in [0,1]$, which determines the priority with which the GCS should be looking at a specific gaze target. The default priority is always 0. 
The Gaze Planner maintains a Gaze Plan ($GP$) which is used by the Gaze Controller to decide which gaze target to look at over a duration of time into the future. The GP stores the priority  $\mathcal{P}_{i,j}$ for each target $\mathcal{T}_i$ at each future time frame  $j$, with $j=0$ being the immediate next time step in the future. We use a time resolution of 200ms for the plan, i.e., each time frame is 200ms long. 

As can  be seen in Algorithm \ref{alg:gazeControlSystem}, at each time step, the Gaze Planner listens to events and updates the relevant $\mathcal{P}_{i,j}$ values, as will be described in detail in the following sections. The relative priorities and specific durations chosen in this paper have either been obtained from literature or iteratively obtained after running the architecture for different scenarios. Thus, we acknowledge that these parameters are somewhat arbitrary, and that more optimal values can very likely be found. It is also possible to generate different robot gaze behaviors (e.g. introvert vs. extrovert) by tweaking the parameters. We leave this for future work. 

At each time step, the Gaze Controller summarizes the current Gaze Plan, $GP_c$, by  calculating the list of final gaze targets for the next 2 seconds into the future (10 time steps). The final gaze targets ($\mathcal{T}_{f,j}$) are calculated as the target that has the highest priority value in each frame $j$ of the $GP$:
\begin{equation} \label{finalGazeCalculation}
    \mathcal{T}_{f,j} = \mathcal{T}_{n}, \underset{n}{\arg\max}(\mathcal{P}_{n,j})
\end{equation}

The current gaze target, $\mathcal{T}_c$, is then equal to the immediate next final target, $\mathcal{T}_{f,0}$. As will be described in Section~\ref{sec:eyeHeadCoordination}, the rest of the $GP_c$ is used to calculate the $slack$ value for the head movement, to achieve natural eye-head coordination. 


After the Gaze Controller has executed the gaze and head movements for that time step, the $GP$ is shifted one step, so that $\mathcal{P}_{i,j} \leftarrow \mathcal{P}_{i,j+1}$.
Since the updates to the $GP$ are done at each time step, it is possible for the Gaze Planner to overwrite $\mathcal{P}$ values in the plan, which makes it possible to dynamically update the plan depending on the events that occur during an interaction.

\begin{figure*}[htp] 
\centering
  \includegraphics[width=\linewidth, height = 10cm]{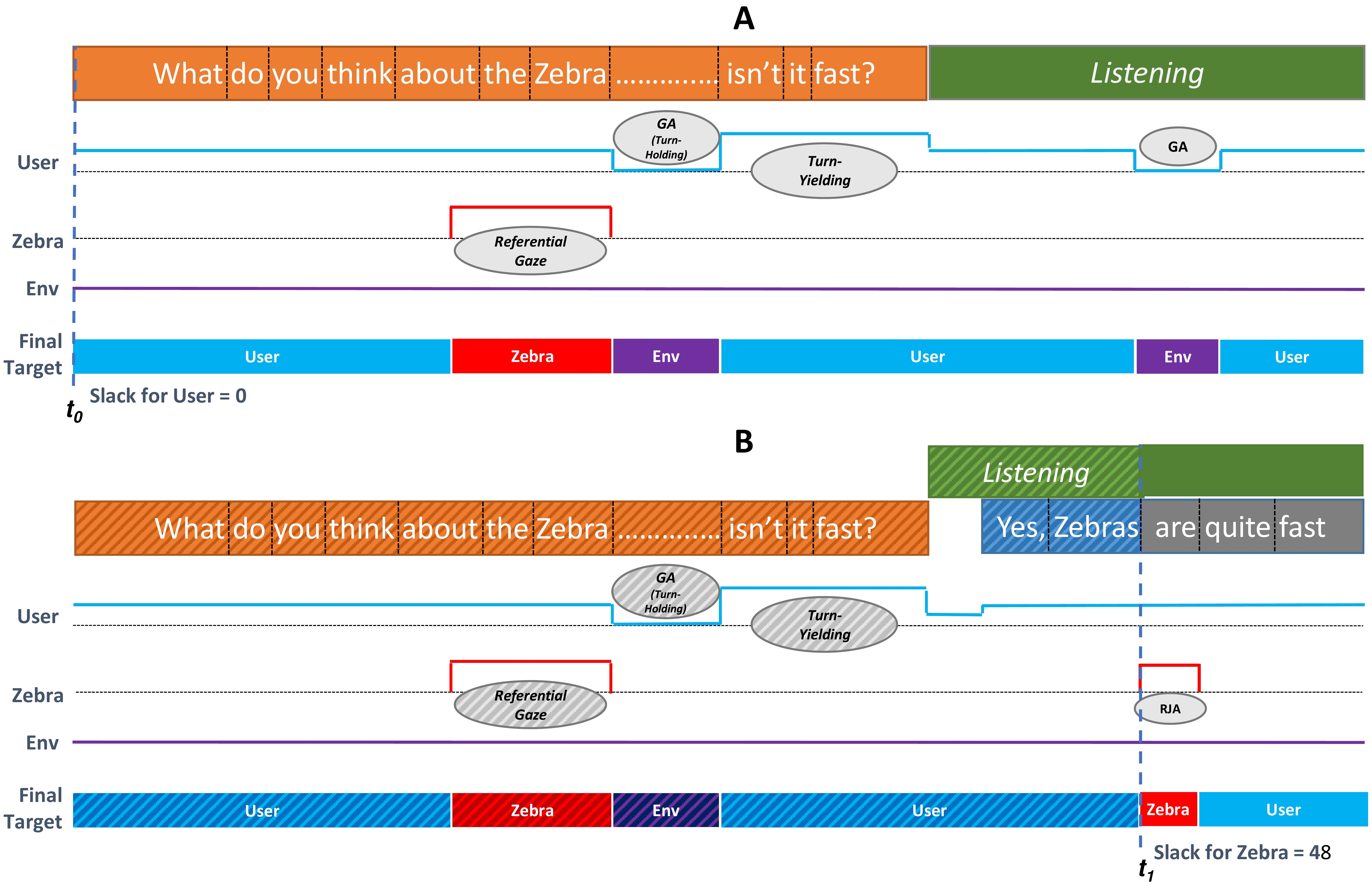}
  \caption{An example of gaze planning done by our GCS.\textbf{(A)} shows the plan at onset of the utterance $t_0$ and \textbf{(B)} shows the updated plan at time step $t_1$. The shaded parts show the already executed plan and the non-shaded part show the plan at the current time step. Grey speech boxes denote that the event is yet to take place.}\label{fig:2}
  \label{gazePlanOutput}
\end{figure*}

Fig.~\ref{gazePlanOutput} shows an example of how the GCS works. Let us consider a scenario where a single user is playing the  ``Animal speed'' Card Game, and a card with Zebra is being discussed. The $\mathcal{P}$ for each $\mathcal{T}$ in the game is plotted below the speech boxes. ``Final Target'' shows the calculated $\mathcal{T}_f$ for the entire example interaction. In the following subsections, we will use this example to discuss the various components of our GCS in detail. 

\subsection{Environment} \label{sec:environment}
As can be seen in Fig.~\ref{gazePlanOutput}, when the $\mathcal{P}$ value for all gaze targets in the plan are 0, the final gaze target is defaulted to the \textsc{environment} (Env). Thus, when the user is given a low priority (e.g., due to gaze aversion) and no other target is given priority, the robot will gaze away from the user. 
\cite{andrist2014conversational} found that the distribution of eye gaze at various regions in the environment depended on the type of gaze aversion being performed. However, to keep the model simple, we randomly select a location in the area around the currently addressed user's face as the location for the \textsc{environment} gaze target.

\subsection{Turn-taking} \label{sec:planning-turn-taking}
At the onset of a robot utterance, the Gaze Planner receives an event from the TTS system which gives information about the entire utterance text, as well as the phonetic transcription with precise timing information. This information can be used to plan the robot's gaze behavior related to speech production and turn-taking (as well as referential gaze, as described in the next section). During the course of the utterance (the $RobotSpeaking$ event), the $\mathcal{P}$ values of the currently addressed users are set to 0.3 which results in the robot looking at the user during the utterance, in the absence of other gaze targets with a higher $\mathcal{P}$ value. This can be seen at the beginning of the Gaze Plan ($t_0$) in Fig.~\ref{gazePlanOutput}A.
This emulates \textbf{mutual gaze}/ \textbf{individual gaze} behavior where the speaker looks at the listener \cite{admoni2017social}.

Speakers tend to avert their gaze to signal that they are thinking or will hold the conversational floor \cite{andrist2014conversational}. The Gaze Planner calculates pause durations in the utterances it is about to speak (using the phoneme timings). If the pause duration is greater than 800ms, the $\mathcal{P}$ of the addressed users are set to 0 for that duration and the \textsc{environment} becomes the $\mathcal{T}_f$ resulting in a turn-holding Gaze Aversion as seen in Fig.~\ref{gazePlanOutput}A. 

By default, the robot will always hold the floor unless the $yielding$ flag in the $RobotSpeaking$ event is set to \textsc{true}. This can be controlled through the Skill API (per default, $yielding$ is set to \textsc{true} in case of a question). When $yielding$ is set to \textsc{true}, the $\mathcal{P}$ values of the currently addressed user targets are set to 0.9, 1000ms before the end of the utterance. This results in a \textbf{turn-yielding} gaze cue \cite{admoni2017social} as can be seen in Fig.~\ref{gazePlanOutput}A when the robot is asking a question. In case $yielding$ is set to \textsc{false}, the $\mathcal{P}$ of the addressed users are set to 0 about 2000ms before the end of the utterance where we do not want the user to barge-in, and gaze aversion is a clear \textbf{turn-holding} cue \cite{jokinen2013gaze}. 

When the robot is not speaking and instead listening to the user (the $RobotListening$ event), the $\mathcal{P}$ of the addressed users are set to 0.4 for the duration of the listening event.
This enables the robot to keep looking at the user unless there is some higher-priority target. When a user starts to speak (the $UserSpeaking$ event), the $\mathcal{P}$ of that specific user target is increased to 0.6. In a multi-party setting, the array microphone of the robot can be used to sense the speech direction, and thereby attribute the speech onset to the right user. This helps in directing the gaze of the robot to the active speaker and is in line with the findings in \cite{vertegaal1999gaze}, where it was found that the listeners always tend to spend the most time looking at the current speaker in multi-party settings.  

\subsection{Joint Attention} \label{sec:jointAttention}
In the Card Game skill, the locations of the cards on the touchscreen and their order are being tracked as task object gaze targets. The $\mathcal{P}$ of the task objects can be raised if Furhat or a user talks about an object, or engages in joint attention in some other way. We do a keyword matching at the onset of the robot utterance (the $RobotSpeaking$ event), to identify any references to a task object. If so, the $\mathcal{P}$ value of that task object is set to 0.9, 1000ms before the specific word is supposed to be spoken (timings are obtained from the TTS system) in order to generate \textbf{referential gaze}. This corresponds to the finding from studies of human communication, where gaze is directed at the referent about 800-1000ms before the reference is made \cite{mehlmann2014exploring}. The same can be seen in Fig.~\ref{gazePlanOutput}A, when the robot refers the Zebra card.

When a task object gaze target (i.e., a card on the touchscreen) is moved, the Gaze Planner sets gaze target's $\mathcal{P}$ to 1 for 2000ms which results in a \textbf{responsive joint attention (RJA)} and Furhat's gaze follows the card that is being moved. The current system is not capable of identifying and tracking objects other than the touchscreen locations. When Furhat is listening to user speech (the $UserSpeaking$ event), we also do a keyword matching on the continuous ASR output to check for any references to the task objects. If a match is found, the corresponding task object's $\mathcal{P}$ is set to 0.7 (for a period of 800ms) after 200ms of the reference being heard. This is in line with what was reported in \cite{mehlmann2014exploring} and is also a form of RJA. This can be seen in Fig.~\ref{gazePlanOutput}B at $t_1$, when user refers the Zebra card.

\subsection{Intimacy Regulation} \label{sec:intimacyRegulation}
Studies have shown that the preferred mutual gaze duration in interactions is between 3-5 seconds before the interlocutor starts to feel uncomfortable \cite{binetti2016pupil}. To avoid this, the  Gaze Planner also takes care of \textbf{intimacy regulating} gaze aversion. At every time step (200ms), the Gaze Planner checks the $GP$ and makes sure that the $\mathcal{T}_c$ is not assigned to a specific user for a duration longer than 3-5 seconds. In example Fig.~\ref{gazePlanOutput}A there is a small period of GA inserted when the planned final gaze target was User for a long duration when listening to the user, which results in intimacy regulating gaze aversions.

\subsection{Eye-Head Coordination} \label{sec:eyeHeadCoordination}
While most of the previous works (see section~\ref{relatedWorks}) have made use of the target angle to coordinate the eye and head movements during gaze shifts, we propose that the planned duration of gaze also plays a role in determining the head and eye movements. If an agent knows that the gaze is planned to be directed at a specific location for a longer period of time, then it can move both the eyes and the head towards that target immediately. On the other hand, if the planned gaze duration is very short, the gaze shift should be done using only the eyes (or with little head rotation). Otherwise, rapidly shifting gaze targets could result in jerky head movements. We can mitigate this problem thanks to the planning approach of our GCS. Additionally, this allows the GCS to take the robot's intention into account when planning the gaze behavior. To the best of our knowledge, this has not been addressed in any previous GCS. 

In our GCS, we introduce a control variable named $slack$ which is the angle by which the head direction is allowed to deviate from the eye gaze direction. As seen in Algorithm \ref{alg:gazeControlSystem}, at every time step (200ms), the Gaze Controller calculates the final gaze targets ($\mathcal{T}_f$) for 2 seconds into the future and summarises them in a list $GP_c$. The current gaze target ($\mathcal{T}_c$) is the first element in the $GP_c$. $slack$ calculation uses the frequency of the same gaze target in the final gaze plan as per equation:
\begin{equation}
    slack = max(48 - (sameTargetFreq * 6), 0)
\end{equation}

$sameTargetFreq$ is the duration of time the gaze is to be directed at a specific target. It is calculated as the frequency of having the same target as the $\mathcal{T}_f$ within a future window of 2s. This determines whether the gaze should be directed using just the eyes or using both the eyes and the head. For example, as in Fig.~\ref{gazePlanOutput}A, it can be seen that at $t_0$, the final target is planned to be User for a long duration. Thus, the $slack$ value is set to 0, and the head direction is fully aligned with the eyes. In Fig.~\ref{gazePlanOutput}B, at $t_1$, the gaze duration planned at the Zebra card is very short so the $slack$ value is set to 48. This means that only eye gaze is directed at the Zebra card for a quick glance. The value 48 has been iteratively obtained.

\begin{figure}[htp] 
  \centering
  \includegraphics[width=\linewidth]{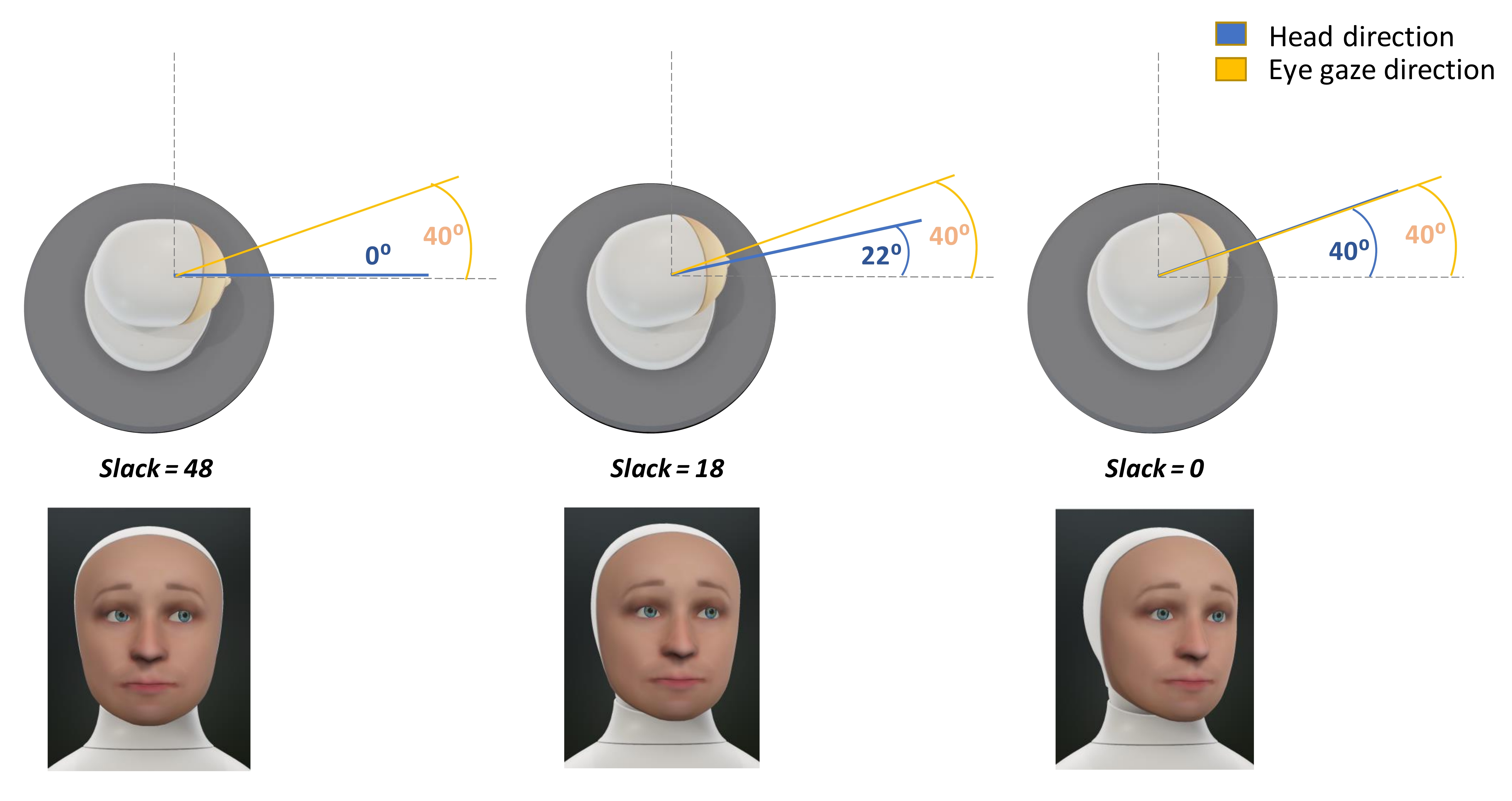}
  \caption{An example showing the differences in eye and head movements when looking at a target depending on the $slack$ values.}
  \label{slackExample}
\end{figure}


Fig.~\ref{slackExample} shows an example of how $slack$ is used in coordinating the eye and head movements towards the gaze target. 
Since the neck movement takes some time, and the eye movement is instantaneous, the Gaze Controller first moves the eyes towards the gaze target and then the neck, while centering the eyes. 

Sometimes, several gaze targets may have an equally high $\mathcal{P}$ value. For example,
in multi-party interactions, two users might be addressees and would have an equally high $\mathcal{P}$ while the robot is speaking. In such cases, it is natural for the speaker's head to be directed somewhere in the middle of the gaze targets, while the eye gaze shifts rapidly between each of them \cite{uemura1980eye,stahl1999amplitude}. We model this behavior in our GCS by checking for any rapid shifts between two or more targets for a future time window. If so, then at every time step, the mid point between the $\mathcal{T}_c$ and the next $\mathcal{T}_f$ is calculated and head is directed at that point.

\section{Experimental Evaluation}
In order to evaluate if a GCS that takes the robot's intention (future gaze behavior) into account is perceived as better, we performed a user study to compare our GCS's performance to a purely reactive GCS.

\subsection{Experimental Setup \& Procedure}
We used the Card Game scenario described in section~\ref{cardGame} for the experiments. A camera was placed behind the participants so that it only captured Furhat's face and the touch screen.

There were two scenarios under which the participants played the Card Game:
\begin{itemize}
    \item \textbf{Planned :} Our GCS with planning.
    \item \textbf{Reactive :} A purely reactive heuristic GCS similar to \cite{pereira2019responsive} was implemented and used as a baseline for the comparative study. The gaze targets were chosen in response to the events taking place in the Card Game (e.g., when the cards were moved, when someone was speaking, etc.). The gaze of the robot was fixed on one target for a duration of 1-5 seconds (same as the original work) before moving to the next target.
\end{itemize}

Each session lasted approximately 30 minutes, during which 2 participants played 2 games (1 from each scenario) together with Furhat. The experiments followed a within-subjects design and the order of scenarios were alternated between sessions. The participants were first guided to their seats in front of the touchscreen and provided with a consent form. Then the researcher briefed the participants about the goal of the study and the way the experiment was going to be conducted. They were told that they would play two games with different versions of the system, but the nature of these two versions was not explained to them, and they were only referred to as scenario 1 and 2.  Participants were encouraged to go through the questionnaire to have a better grasp of what to look out for before starting the first game. After each game, the participants filled out one part of the questionnaire. The questionnaire had two 9-point likert scales placed under each question; one for each scenario. The participants were asked to score the questions based on how they perceived the interaction in terms of the question being asked. They were asked to look for differences in the scenarios and make different judgements where applicable.
The participants were instructed not to discuss the scenarios with each other before filling out the questionnaire. At the end of the session, the participants were also asked to choose which of the two interactions they preferred. 

\subsection{Data Collection and Evaluation}
We recruited 28 participants to take part in the user study (14 males and 14 females) with ages ranging between 18 and 51 (mean = 32.92, SD = 8.22), and participants were paired up. The responses from 2 participants were removed from the analysis, as they violated the instructions and discussed the scenarios with each other prior to filling out the questionnaire. No prior interaction with social robots was needed before participating in the experiment. The experiments were conducted in English.

The questionnaire had 10 9-point Likert scale questions which were grouped into 5 dimensions, as can be seen in Table~\ref{tab:questionnaire}. As the goal of our study was to to compare two GCSs, we selected the dimensions and questions based on aspects that should be important for a good GCS. For each dimension, the mean score of the responses to the individual questions in that dimension was calculated. We refer to this as the \textit{dimension score}. For a better GCS, we expected the \textit{dimension scores} to be high for all dimensions, except for the \textit{Intimacy} dimension, which should be lower, given how the questions were formulated. We used the statement “Furhat kept staring at me too much” as a sign of bad Intimacy regulation, since periodic GA while listening leads to making speakers more comfortable and reduces negative perceptions~\cite{abele1986functions}. While we use the term Intimacy as a short label for this dimension, this question does not of course capture all aspects of intimacy, but it was designed to be easy to interpret for the participants. 

At the end of the experiment, the participants were also asked (verbally) which scenario they preferred, taking all factors into account. 



\begin{table}[htbp]
\centering
  \caption{Questionnaire used for evaluation}   \label{tab:questionnaire}
  \renewcommand{\arraystretch}{1.1}
  \vspace{-0.6cm}
  \begin{center}
   \begin{tabular}{p{1.5cm} p{6.4cm}}
     \textbf{Dimension} & \textbf{Question}\\
    \hline
      & Furhat looked at the cards at the right time.\\
     Awareness & Furhat was aware of what was happening in the game.\\
     \hline
      & I could interpret Furhat's intention from its gaze.\\
     Interpretation & Furhat's gaze helped me understand its instructions better.\\
     \hline
      & I was able to understand when Furhat wanted me to speak.\\
     Turn-taking & I was able to understand when Furhat wanted to keep speaking.\\
     & I was able to understand when Furhat was talking to me.\\
     \hline
      & Furhat's gaze was human-like.\\
     Human-likeness & The coordination between eye and head movements seemed natural.\\
     \hline
     Intimacy & Furhat kept staring at me too much.\\
  \hline
\end{tabular}
 \end{center}
\end{table} 
\vspace{-0.6cm}
\section{Results}
\begin{figure}[htbp] 
  \centering
  \includegraphics[width=\linewidth]{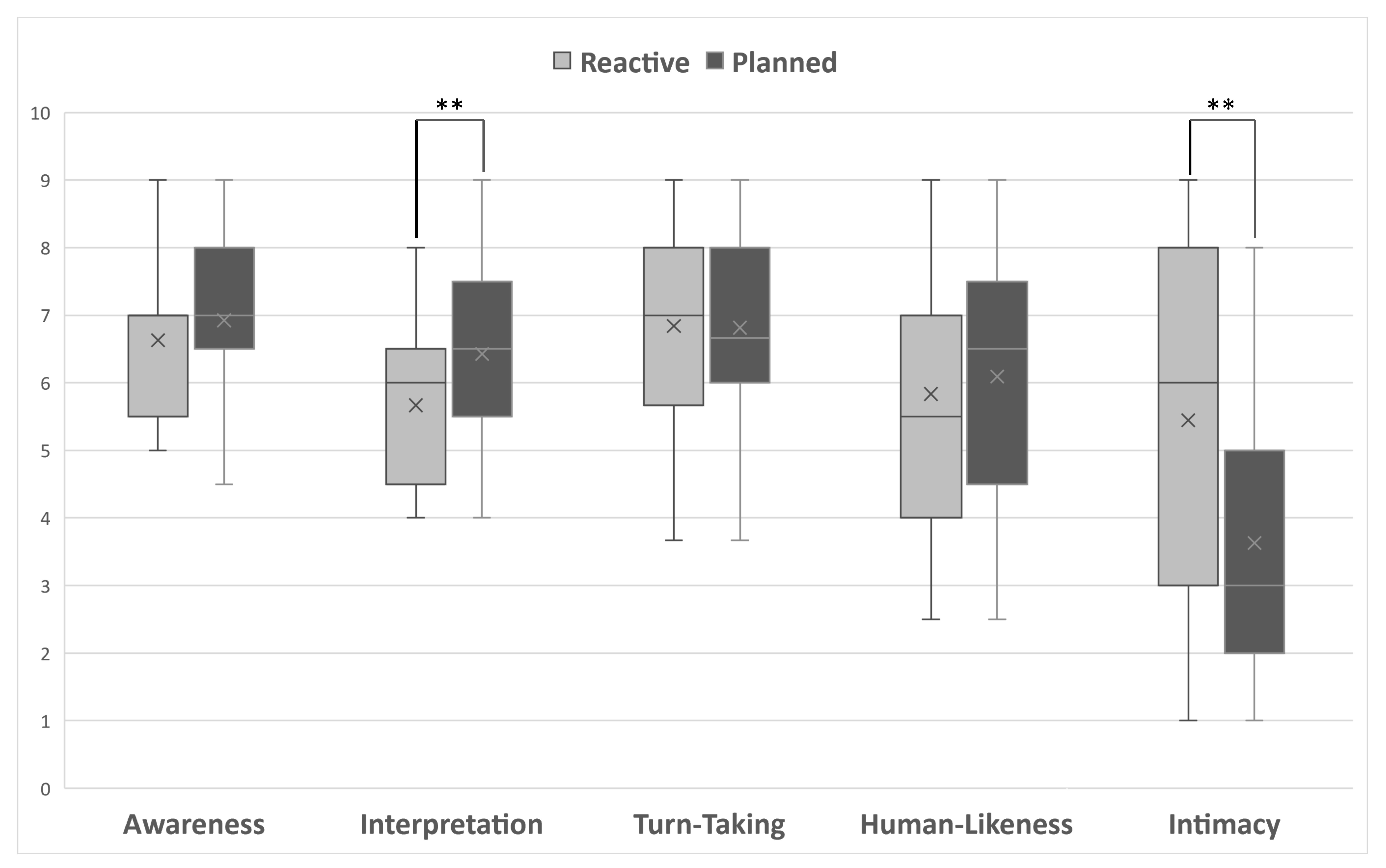}
  \caption{The comparison between the responses obtained for the reactive system and the proposed GCS. The grey bar in the box plot indicates the median, the $\times$ denotes the mean, the boxes show the upper and lower quartiles of the data. The bars on both ends of the vertical lines denote the maximum and minimum values in the data. ** ($p < 0.01$)}
  \label{resultsImage}
\end{figure}

Fig.~\ref{resultsImage} shows the \textit{dimension scores} of each dimension for both GCS versions, based on the responses from the 26 participants. As our hypothesis was that they would prefer the Planned version, we performed a one-tailed Wilcoxon signed-rank test. Since we compared five dimensions, we set $\alpha=0.01$, after Bonferroni correction. Significant results were obtained for \textit{Interpretation} (p = 0.007) and \textit{Intimacy} (p = 0.0013). While the mean values for \textit{Awareness} and \textit{Human-Likeness} were higher for the Planned version, the differences were not statistically significant. 

For the final preference question, 19 participants preferred the Planned version, whereas 4 preferred the Reactive version, and 4 could not decide. We found the results to be significat with p = 0.0002 and $\chi^2$ = 16.66. 
\section{Discussion}
The goal of our evaluation was to do a comparison between a purely reactive GCS and the planning-based GCS proposed in this paper. We hypothesised that using a GCS with planning can lead to improved perception of the robot's gaze behavior during the interaction. The results from the evaluation indicate that our GCS was significantly more interpretable, had better intimacy regulation, and was generally preferred over the reactive version. 

When it comes to the dimensions of Awareness, Turn-taking and Human-likeness, we did not find any significant differences. One possible explanation could be that it might have been difficult for the participants to observe subtle gaze behaviors while being engaged in playing a new game.
Previous studies on turn-taking models for conversational systems have shown that it is very hard for the participants to judge subtle things like turn-taking while interacting themselves \cite{Meena2014}.
This is also in line with the \textit{Load Theory} \cite{lavie2004load}, which says that when there is a higher cognitive load, the selective attention performance becomes poor. 
Another problem that was noticed during the experiments was that the sound source localization on the robot was not always working very well. This means that Furhat sometimes turned to the wrong participant, which clearly could have affected the perception of the Turn-taking dimension. 

Another issue could be the novelty effect; most of the participants were interacting with a social robot for the first time. Attention might have been split in familiarizing themselves with the robot and its capabilities during the first scenario, which could have impacted the ratings. A potential way of mitigating this problem in future studies could be to let the participants first do a test round where they familiarize themselves with the robot. 
A potential follow-up study could be to show the recordings of the experiments to third-party observers and let them compare the two versions. In doing so, the participants would be able to solely focus on rating the robot's gaze during the interactions. \cite{Meena2014} reported that third-party observers could easily perceive differences between different turn-taking models, unlike the participants who were engaged in the interaction. 

It should also be noted that many of the parameters chosen for the model could of course be further tuned. We can also envision a hybrid model, where certain priority scores are being data-driven, and others are rule-driven. 

\section{Conclusion}
In this paper, we proposed and implemented a novel planning-based comprehensive GCS to automate the gaze behavior of social robots. The system is capable of planning the gaze behavior for a future rolling time window of fixed length, and use this plan to coordinate the eye and head movements of the robot taking the robot's intention into account. We conducted a user study to evaluate our GCS and compared it with a purely reactive GCS. The results suggest that a GCS with such type of planning is perceived to be significantly more interpretable and has better intimacy regulation. It was also found that overall, our GCS was preferred over the reactive system when users were asked to choose one. This shows that planning is an important aspect of gaze control, which has not been considered in previous works.

\addtolength{\textheight}{0cm}   


\bibliographystyle{IEEEtran}
\bibliography{IEEEabrv,paper}

\end{document}

%% file: gazeModels.tex
\begin{table*}[htp]
\centering
  \caption{Review of Gaze Models in HRI. 
  }
  \label{tab:gazeModels}
     \renewcommand{\arraystretch}{1.15}
   \begin{tabular}{c|c|c|c|c|c|c|c}
     \hline
     Paper & \multicolumn{1}{l|}{Multi-party} & Modelling & Planning & Gaze Aversion & RJA\footnote{} & Referential Gaze & Turn-Taking\\
     \hline
     Mutlu et al. 2012\cite{mutlu2012conversational} & \multicolumn{1}{l|}{Yes} & Data-driven & No & No & No & No & Yes\\
     Andrist et al. 2014\cite{andrist2014conversational} & \multicolumn{1}{l|}{No} & Data-driven & No & Yes & No & No & Yes\\
     Mehlmann et al. 2014\cite{mehlmann2014exploring} & \multicolumn{1}{l|}{No} & Heuristic & No & Yes & Yes & Yes & Yes\\
     Zaraki et al. 2014\cite{zaraki2014designing} & \multicolumn{1}{l|}{Yes} & Heuristic & No & No & No & No & Yes\\
     Andrist et al. 2015\cite{andrist2015look} & \multicolumn{1}{l|}{No} & Data-driven & No & No & Yes & No & Yes\\
     Nakano et al. 2015\cite{nakano2015generating} & \multicolumn{1}{l|}{Yes} & Data-driven & No & No & No & No & Yes\\
     Lehmann et al. 2017\cite{lehmann2017naturalistic} & \multicolumn{1}{l|}{No} & Data-driven & No & No & No & No & Yes\\
     Zhang et al. 2017\cite{zhang2017look} & \multicolumn{1}{l|}{No} & Heuristic & No & Yes & No & No & Yes\\
     Zhong et al. 2019\cite{zhong2019investigating} & \multicolumn{1}{l|}{No} & Heuristic & No & Yes & No & No & Yes\\
     Pereira et al. 2019\cite{pereira2019responsive} & \multicolumn{1}{l|}{No} & Heuristic & No & Yes & Yes & Yes & Yes\\
     Lala et al. 2019\cite{lala2019smooth} & \multicolumn{1}{l|}{No} & Heuristic & No & Yes & No & No & Yes\\
     Stefanov et al. 2019\cite{stefanov2019modeling} & \multicolumn{1}{l|}{Yes} & Data-driven & No & No & Yes & Yes & Yes\\
     Pan et al. 2020\cite{pan2020realistic} & \multicolumn{1}{l|}{Yes} & Heuristic & No & No & No & No & Yes\\
     \hline
     Proposed Gaze model & \multicolumn{1}{l|}{Yes} & Heuristic & Yes & Yes & Yes & Yes & Yes\\
     \hline
    \multicolumn{8}{p{15cm}}{\footnotesize \textsuperscript{1} Responsive Joint Attention. This is set to ``Yes'' if the GCS was capable of responding to user's referential gaze, verbal references to objects/ interlocutors/ locations, movement of task object etc.}
\end{tabular}
\end{table*}